\documentclass[runningheads]{llncs}
\usepackage{graphicx}
\usepackage{tikz}
\tikzstyle{every node}=[font=\fontsize{22}{22}\selectfont]
\usepackage{subfiles}
\usepackage[linesnumbered,ruled,vlined]{algorithm2e}
\usepackage{amsmath}
\usepackage{amsfonts}
\usepackage{multirow}
\usepackage{hyperref}
\usepackage{array}
\usepackage{float}
\usepackage{comment}
\usepackage[misc,geometry]{ifsym}

\begin{document}
\title{Fashion Style Generation: \\Evolutionary Search with Gaussian Mixture Models in the Latent Space
}
\titlerunning{Evolutionary Fashion Style Generation with GANs and GMMs}
%

\author{Imke Grabe\inst{1}
\and
Jichen Zhu\inst{1}
\and
Manex Agirrezabal\inst{2}
}
\authorrunning{I. Grabe et al.}
%
\institute{IT University of Copenhagen, Copenhagen, Denmark\\ \email{\{imgr,jicz\}@itu.dk} \and
University of Copenhagen, Copenhagen, Denmark\\ \email{manex.aguirrezabal@hum.ku.dk}\\
}
\maketitle              
\begin{abstract}
This paper presents a novel approach for guiding a Generative Adversarial Network trained on the \textit{FashionGen} dataset to generate designs corresponding to target fashion styles. 
Finding the latent vectors in the generator's latent space that correspond to a style is approached as an evolutionary search problem. 
A Gaussian mixture model is applied to identify fashion styles based on the higher-layer representations of outfits in a clothing-specific attribute prediction model.
Over generations, a genetic algorithm optimizes a population of designs to increase their probability of belonging to one of the Gaussian mixture components or styles. 
Showing that the developed system can generate images of maximum fitness visually resembling certain styles, our approach provides a promising direction to guide the search for style-coherent designs. 

\end{abstract}
\section{Introduction}

In many areas of music, arts, and design, \textit{artificial intelligence} (AI) can procedurally generate new cultural artifacts~\cite{marchetti2021convolutional,elgammal2017can,xin2021object,zhu2013shall,zhu2018explainable}. 
With the recent developments of \textit{Generative Adversarial Networks} (GANs), AI technology can become a powerful asset for human creators to design complex objects. 
For example, researchers have explored how to use GANs to design fashion artifacts in fashion design. The emerging area of {\em intelligent fashion} investigates the detection and recommendation of clothing items, the analysis of style trends, and the synthesis of clothing~\cite{cheng2021fashion}. As part of fashion systhesis, GANs have been applied in the creation of new items~\cite{kato2019gans}, the simulation of try-on scenarios~\cite{yildirim2019generating}, or for personalized design~\cite{zhu2017your}. 
Because fashion plays a fundamental part in human culture, it is essential to investigate how generative AI might contribute to its creation.

This paper focuses on the generation of fashion styles, defined as visual themes, such as the combination of clothing artifacts or attributes as part of an outfit (e.g., blue pants and a vertically striped shirt). 
By extracting the representation of images on higher-level layers of an attribute prediction model, styles can be identified based on high-dimensional visual themes~\cite{matzen2017streetstyle}. 
Existing work in intelligent fashion has analyzed styles~\cite{takagi2017makes,kiapour2014hipster,hsiao2017learning}, ultimately allowing for the prediction of trend behavior~\cite{al-halah2020b}. 
Research on fashion style generation with GANs focuses on controlling specific features~\cite{yildirim2018disentangling,jiang2021deep}, conditioning design with a text encoding ~\cite{zhu2017your}, or transferring an exact outfit to other poses~\cite{zhu2017your,yildirim2019generating}. The control with regards to fashion styles, as defined above, has, however, not been considered in the generative process.

An open problem in generative design is how to guide the generation of GANs towards desirable outcomes. The GANs' generator network learns to map random input variables to the complex output features resembling the training data during the training. 
The procedure creates an entangled latent space, 
making it impossible to inspect how changes in the latent code affect the semantic output features. This entanglement impedes the control of the generated designs towards a desired look.
Differentiated control of the latent features of generative clothing models has been achieved by conditioning the generator with the encoding of a text description in the latent vector~\cite{zhu2017your}, or by disentangling color, texture, and shape inputs through separate losses in the loss function during training~\cite{yildirim2018disentangling}. Approaches like \textit{StyleGAN}~\cite{karras2019style} aim at controlling certain stylistic features in images, such as the transfer of a complete outfit~\cite{yildirim2019generating}. However, guiding the generation of designs towards fashion styles consisting of broader visual themes remains an open research problem. 
Fashion describes the specific category of clothing driven by the developments of style trends. As fashion styles capture meaningful temporal and local developments with societies~\cite{mackinney-valentin2010,al-halah2020b,matzen2017streetstyle}, responding to such themes matters for the generative process.

This paper presents a new method to guide GANs using a \textit{Gaussian mixture model} (GMM). While GMMs have been used in the field of intelligent fashion to identify fashion styles~\cite{matzen2017streetstyle,al-halah2020b}, they have not been applied to support generative purposes. 
To better control the GANs' entangled latent space, we utilize the GMM to find the latent vectors that correspond to \textit{stylistic} designs in an evolutionary search problem. More specifically, our method combines (1) generative deep learning, (2) the analysis of fashion styles, and (3) the application of genetic optimization algorithms to search a design space. 

The main contribution of this work is a new method to guide GANs using a GMM. This paper presents the method proposed in our prior work \cite{grabe2021evolfash}.  
It allows GANs' generative process to be guided based on higher-level themes, instead of separate attributes as in existing approaches~\cite{yildirim2018disentangling,jiang2021deep}. We tested our method in the context of generating fashion styles. We found that while our proposed framework generally supports the generation of designs according to target styles, the GMM-based fitness measure is not always aligned with visual coherency to the styles.
Some generated designs reveal that the fitness measure relies on a machine-specific understanding of style. Further investigation into the interplay between style model and the exploration of the latent space and the parameter setting of the genetic algorithm is required to align the results with a human understanding of style.

The remainder of the paper is organized as follows. After presenting related work, 
we introduce our dataset and proposed model, consisting of a GAN model, a style model, and the evolutionary search connecting the former. Next, we present our experimental results and conclude with discussions.

\section{Related Work}
This section presents related work in the three research areas relevant to this study, namely within (1) GANs, (2) analyzing fashion styles, and (3) applying evolutionary search to steer the generation process.

\subsection{GANs}
The introduction of GANs revolutionized the creation of computer-generated content~\cite{goodfellow2014generative}.
GANs, consisting of two neural networks competing against each other as generator and a discriminator, learn to generate outputs by resembling a training distribution. For example, GANs can generate clothing artifacts when trained with a dataset of those. 
Notably, the training method of \textit{Progressively growing GANs} (P-GANs) supports the generation of high-resolution images, as was demonstrated by Rostamzadeh et al.~\cite{rostamzadeh2018fashion} with their introduction of a fashion dataset.
In fashion generation, different objectives have been guiding the training of GANs, such as conditioning their output with the text description of desired looks~\cite{zhu2017your}, or color, texture, and shape~\cite{yildirim2018disentangling}.

Notice that for GANs, the term {\em style} is typically used to describe the manipulation of a design with regards to specified visual attributes. 
StyleGANs~\cite{karras2019style} provide an architecture that disentangles the latent space of the generator network, allowing for a targeted modification of high-level to low-level attributes corresponding to different resolutions in the network. 
Yildirim et al.~\cite{yildirim2019generating} trained a StyleGAN to transfer a complete outfit to other models, as in a try-on scenario. This notion sets focus on certain visual attributes, similar to Jiang et al.~\cite{jiang2021deep}, who apply GANs to transfer patterns to shirt designs. 

\subsection{Fashion styles}
Studies have addressed the phenomenon of styles in fashion from different angles, varying from weak~\cite{liu2016deepfashion,ge2019deepfashion2} to strong~\cite{takagi2017makes,kiapour2014hipster} style annotations, as well as their unsupervised discovery~\cite{hsiao2017learning,al-halah2020b,matzen2017streetstyle}. 
Drawing on the latter, this paper builds on research that approaches fashion styles as a ``mode in the data capturing a distribution of attributes''~\cite[p.7]{al-halah2020b}. Hence, they can be discovered unsupervised by clustering attribute predictions of images. 
Clustering models such as a Gaussian mixture model (GMM) have previously been applied to find recurring components in the attribute embedding of images~\cite{matzen2017streetstyle,al-halah2020b}.
After identifying styles with a GMM, projecting the embedding of any (generated) image onto the GMM can measure of how well it fits into the discovered styles. In that way, we can evaluate how an image resembles a particular style. 

Instead of using the prediction scores of an attribute prediction model as the embedding, previous layers of the model also capture meaningful  
themes in images.
Matzen et al.~\cite{matzen2017streetstyle} introduce a method that finds styles as Gaussian mixture components in the projections of images onto the model's penultimate feature space. 
The second-last layer captures learned features that are more specific than the output of the final fully-connected layer. It serves as a valuable embedding for recognizing high-level themes beyond attribute scores~\cite{mahmood2020automatic}. 
The resulting clusters can be understood as a ``global visual vocabulary for fashion''~\cite[p.7]{matzen2017streetstyle}. 
Letting this \textit{language of fashion} inform the generative process by GANs is the objective of our suggested framework. 
We adapt Matzen et al.'s~\cite{matzen2017streetstyle} procedure to find styles in the dataset \textit{FashionGen} to use them for guiding the generation towards a chosen style. 

To sum up, as the subject of fashion analysis, \textit{style} refers to the broader sense of visual themes in outfits. Unless otherwise specified, the rest of the paper will use the term \textit{style} to refer to this definition. 
Going beyond the attributes corresponding to the different layers of the GAN, specific items, or textures, the concept has not been considered in generation yet. 
We suggest a system affording the generation of designs based on fashion styles discovered in an unsupervised manner by a GMM. 
Embedded into an evolutionary search, the GMM guides the generation of designs.

\subsection{Evolutionary search of GANs' latent space}
Our project aims to generate designs of various styles with visually diverse attributes with GANs. Navigating the networks' complex latent space with this objective requires changing minor and major visual features.
Drawing on the smooth characteristic of GANs' latent space, evolutionary search can optimize its latent variables~\cite{bontrager2018deepmasterprints}. 
More specifically, a genetic algorithm alters a population of latent vectors by recombining and mutating them~\cite{eiben2003introduction}. Through selection based on a fitness objective, such as maximizing the probability of belonging to a target style, latent vectors improve over many generations.

Previous work has applied genetic algorithms to explore the latent space with different goals. While Roziere et al.~\cite{roziere2020evolgan} improve the quality of only one fashion image generated by a P-GAN, Fernandes et al.~\cite{fernandes2020evolutionary} evolve a set of latent vectors to increase the diversity among the corresponding set of generated images. 
Instead of a pre-defined fitness objective, interactive genetic algorithms use user evaluation as a measure of fitness. Designing smaller clothing items in this interactive manner has been proposed in \textit{DeepIE}~\cite{bontrager2018deep}, further developed into \textit{StyleIE}~\cite{styleIE2021}. 
Where an interactive genetic algorithm uses a human-in-the-loop to assess fitness, we apply a GMM to measure the fitness of a generated image, following the goal of generating designs that belong to an automatically discovered target style.
Our contribution is to create a fully automated method for style generation by adding a GMM to the evolutionary loop. Instead of a human who chooses images of desired properties, the selection is based on the images' projection onto the clustering space.

\section{Dataset}
FashionGen's~\cite{rostamzadeh2018fashion} clothing partition in a resolution of $256\times256$ pixels is used as a dataset for training the generative model and the style model. Consisting of over 200,000 images of outfits worn by a model, each outfit is represented in four different poses in the subset. 
The dataset mainly consists of images of whole-body outfits taken under consistent lighting conditions in front of white backgrounds, providing the ideal conditions for generation and clustering to focus purely on fashion attributes and identify styles across a combination of artifacts.

\section{Model}
\begin{figure}[tb]
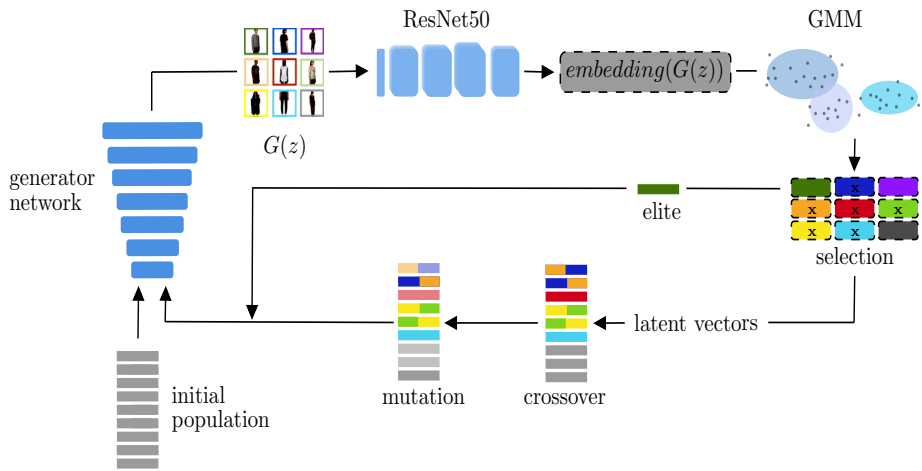
 
\centering
\tikzset{every picture/.style={line width=0.75pt}}
\subfile{diagrams/evol_diagram.tex}
 \caption[Framework for searching the generative model's latent space with the help of the clustering model.]{Model of the genetic algorithm for searching the generative model’s latent space with the help of the clustering model.
 First, the trained generator network creates images based on randomly initialized latent vectors.
 Next, the generated images are represented as their embedding of the ResNet50 and then projected onto the clustering space of the GMM.
 The fittest ones are selected based on their posterior probability of belonging to the target cluster.
 Additionally, the elite, is preserved for the next generation.
 The others are recombined and new individuals are added before being mutated. The resulting latent vectors lay the basis for the next generation of images.}
\label{tikzmodel-ev}
\end{figure}

Our proposed framework consists of three parts: A generative model, a style model, and the evolutionary search connecting the two. 
To make the (1) generative model output designs of a style that is discovered by the (2) style model,
parts of the two models are combined in an (3) evolutionary search algorithm, as illustrated in Fig. \ref{tikzmodel-ev}. The details of the components are presented in this order.\footnote{The code is available:  \href{https://github.com/imkegrabe/fashionstyle-generation-GMM}{https://github.com/imkegrabe/fashionstyle-generation-GMM}}

\subsection{Generative model} \label{methgan}

As the first part of the framework, a GAN is trained on the complete dataset. 
The P-GAN architecture and training procedure are utilized, which applies the \textit{Wasserstein GAN} loss with gradient penalty (WGAN-GP)~\cite{karras2017progressive}.\footnote{We use the implementation made available by Facebook Research: \href{https://github.com/facebookresearch/pytorch_GAN_zoo}{https://github.com/facebookresearch/pytorch\_GAN\_zoo}}
Hence, the goal of the training procedure is to minimize the following function with respect to $G$, and maximize it with respect to $D$:
\begin{equation}
    \begin{aligned}
    \min_G\max_D \mathbb{E}_{x\sim p_{data}}[D(x)] - \mathbb{E}_{z \sim p_{z}}[D(G(z))]
    + \lambda \mathbb{E}_{\hat{x}\sim p(\hat{x})} [(|| \nabla_{\hat{x}} D(\hat{x})||_2 -1)^2]
    \end{aligned}
\end{equation}

By competing against a discriminator network $D$, a generator network $G$ learns how to map an input vector $z$ of 512 variables,\footnote{The implementation expands the latent codes with 20 additional variables based on the outfits’ item representation. As initial experimentation did not make their effect on the evolutionary search behaviour apparent, we treated them as the other latent variables. Their role should be further examined in future experiments.}
randomly sampled from a standard normal distribution $\mathcal{N}(\mu=0,\sigma=1)$, to output images resembling the real data distribution in the dataset. 
The model was trained on a \textit{NVIDIA Tesla V100-SXM2-16GB} GPU for seven days to create images of resolution $256\times256$. Table \ref{tab:ganparams} provides an overview of the training parameters.

\begin{table}[tb]
\centering
\caption{GAN training parameters. Note that the batch size was changed to 8 for the last training epoch.}
\centering
\begin{tabular}{ p{4cm}>{\raggedleft\arraybackslash}p{4cm}}
    \hline
    Parameter & Setting \\
    \hline
    Optimizer & Adam\\
    Activation function & leakyRELU \\
    Learning rate &  Equalized learning rate\\
    Batch size & 16 (8) \\
    Loss function & WGAN-GP\\
    Noise distribution & $\mathcal{N}(\mu = 0, \sigma = 1)$\\
    \hline
\end{tabular}
\label{tab:ganparams}
\end{table}


\subsection{Style model} \label{methstyle}
The goal of the second part of the framework is to discover styles in the dataset. Informed by the method in~\cite{matzen2017streetstyle}, the visual embedding learned by an attribute prediction model is leveraged to cluster outfits into styles.

We create image representations using a \textit{ResNet50}.\footnote{The backbone of the attribute prediction model provided by the open-source toolbox for visual fashion analysis by the Multimedia Lab, Chinese University of Hong Kong is adapted: \hyperlink{https://github.com/open-mmlab/mmfashion}{https://github.com/open-mmlab/mmfashion}} 
The network was pre-trained to map input images of \textit{DeepFashion} to 1000 clothing-specific attributes instead of more general ones as in models pre-trained on ImageNet. The pre-training makes it a robust feature basis for analyzing datasets containing a wide range of clothing items.

\begin{figure}[tb]
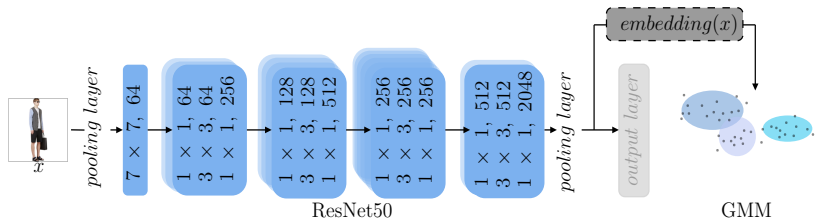

\centering
\subfile{diagrams/style-model.tex}
 \caption[Style model consisting of an attribute prediction model and a clustering model.]{Style model consisting of a ResNet50 as feature extraction model and a GMM as clustering model. The ResNet consists of several units containing 1-6 layer(s). A layer contains convolution(s) in the form of kernels, here notated as $r \times r, n$ with kernel size $r$ and $n$ channels.
 The output layer is discarded to extract the embedding for image $x$ from the pooling layer. Based on the embedding for all outfits in the dataset, $k$ Gaussian mixture components are determined.
 }
  \label{attrstylemodel}
\end{figure}

As we are concerned with finding subtle styles, 
we approach them as a visual concept situated between pre-defined coarse categories and low-level single-feature attribution.
To discover coherent style clusters representing these subtle visual themes, the penultimate layer of the ResNet50 serves as a meaningful embedding space. 
The last layer before the linear layer outputting the prediction scores for attributes captures high-level features of fashion images, though not directly class-specific to DeepFashion.
In practice, the output of the last convolutional unit, more precisely after it has been converted to a 2048-dimensional vector by the pooling layer, is extracted as the $embedding(x)$ of image $x$, as can be seen in Fig. \ref{attrstylemodel}.

Based on the embedding, clustering is employed to find recurring themes in the embedding space.
To capture the distribution of all outfits in the style model while at the same time making the clustering computationally efficient, 
we chose only images in pose 4, which show most full-body images, chosen to build the style model.
Hence, an embedding is retrieved for one image per outfit in the dataset.
The embedding vectors are scaled to zero mean before principal component analysis (PCA) projects them onto the 135 principal components capturing 90\% of the embedding's variance.

A GMM is applied to find a mixture of Gaussian probability distributions that represent the data distribution. 
Through experimentation in line with Matzen et al.~\cite{matzen2017streetstyle}, we find that $150$ mixture components seem to capture visually coherent fashion styles in the dataset. These style clusters range from 60 to 1100 ($\mu = 314.2$) images in size. Any image's posterior probability $p_t$ of belonging to a component $t$ depicts how well it represents a style.
The probabilistic model guarantees that increasing an image's posterior probability of belonging to a cluster component reduces its probability of belonging to other clusters.

\subsection{Evolutionary search}

The trained generator of the GAN acts as a genotype-to-phenotype mapping, 
where the latent encoding, interpreted as the genotype, governs the appearance of the output designs, the phenotype.
Adjusting the genotypes should move the corresponding phenotypes projected onto the GMM closer to a target style. 
Over several generations, the proposed genetic algorithm alters a population of $N_{pop}$ latent vectors initialized from the distribution as the latent variables. The algorithm selects latent vectors for the next generation by evaluating their probability of falling into the targeted style cluster. The goal is to arrive at a set of latent vectors representing designs of the desired style.
While the following sections explain the details, Algorithm 1 shows the pseudo-code for the procedure.\footnote{The genetic algorithm was implemented using the
evolutionary computation framework 
DEAP: \href{https://deap.readthedocs.io/en/master/index.html}{https://deap.readthedocs.io/en/master/index.html}}
\begin{algorithm}[!ht]

\SetAlgoLined
\DontPrintSemicolon

\KwResult{$G(z^*)$ closest to style cluster centroids}
  \SetKwFunction{fitness}{fitness}
  \SetKwProg{Fn}{Function}{:}{}
  \Fn{\fitness{z, t}}{
  \KwRet $p_t(embedding(G(z)))$ \;
  }
  
\BlankLine
  \SetKwFunction{selection}{selection}
  \SetKwProg{Fn}{Function}{:}{}
  \Fn{\selection{population, N}}{
    \For{$i \gets 1$ \text{ to } $N$}{
        $best_i \gets$ winner out of $N_{ts}$ randomly chosen $z$ with \fitness{z,t}\;
        
        }
   \KwRet $best_1, ... , best_N$ \;
  }
  
\BlankLine
  \SetKwFunction{crossover}{crossover}
  \SetKwProg{Fn}{Function}{:}{}
  \Fn{\crossover{a, b}}{
        \For{i in $a$}{
        $a_i = \alpha a_i + (1-\alpha) b_i$ for $\alpha \sim Bernoulli(0.5)$\;
        $b_i = \alpha b_i + (1-\alpha) a_i$ for $\alpha \sim Bernoulli(0.5)$\;
        }
        \KwRet $a, b$ \;
  }
  
\BlankLine
  \SetKwFunction{mutation}{mutation}
  \SetKwProg{Fn}{Function}{:}{}
  \Fn{\mutation{a}}{
        $noise \gets $ vector of length $l$ where $noise_i \sim \mathcal{N}(\mu=0, \sigma=1)$\;
        \KwRet $a + noise$ \;
  }

\BlankLine
$population \gets N_{pop} \times z$\;
\For{g in $N_{gen}$}{
 $elite \gets N_{elite} \times z$ with maximum \fitness(z,t) in $population$\;
 $population =$ \selection{population, $N_{pop}$} \;
 
 \For{$a, b$ in population}{
        \If{random $<p_{cx}$}{
        $a, b$ = \crossover{$a, b$}
        }
        }
 
 $population = population + N_{new} \times z$\;
 
 \For{$a$ in $population$}{
\If{$random < p_{mut}$}{
   $a$ = \mutation{a}\;
   }

 }
 $population = population + elite$\;

   }
   
 \caption{Evolutionary Search of GANs' latent space using GMM.}
 \label{evolalg}
\end{algorithm}
\subsubsection{Representation and transformation}
A generated design is represented by its latent vector 
$
    z = \langle v_1,...,v_l\rangle\text{ with }v\in\mathbb{R}
$
consisting of $l$ latent variables.

While an initial latent vector $z$ is sampled randomly from the underlying distribution (see section \ref{methgan}),
the genetic algorithm aims to transform $z$ towards $z^*$ with $G(z^*)$ representing a design of the targeted style. This transformation is guided by the fitness objective defined below.

\subsubsection{Fitness and selection}
To arrive at the desired goal, individuals are evaluated against a fitness measure. 
Recall from section \ref{methstyle}, that an image's posterior probability of belonging to a style component $t$ is $p_t$. 
Following the goal of resembling a certain target style $t$, the fitness criterion $\mathcal{F}_t$ is defined by an individual's posterior probability of belonging to the respective target style cluster, referred to as $f_t$. Applying the measure brought forward by the GMM, $f_t = p_t$.

To assess the fitness of a latent vector, image $G(z)$ is generated.
The generated image is then projected onto the embedding space to retrieve $embedding(G(z))$. 
Finally, the $embedding(G(z))$ is projected onto the GMM, where the probability of belonging to target cluster $t$ is extracted as $p_t$, representing the fitness criterion $f_t$. That defines the fitness $f$ of an individual, or latent vector $z$, of belonging to a target cluster $t$ as 
\begin{equation}
    f_t(z) = p_t(embedding(G(z))).
\end{equation}

For the generated images to fit into a style cluster, the fitness $\mathcal{F}_t$ needs to be maximized to obtain a latent vector $z^*$ of target style $t$:
\begin{equation}
\centering
    \begin{aligned}
    z^* = \arg \underset{z}{\max} \text{ } \mathcal{F}_t(z)
    \end{aligned}
\end{equation}

\noindent
By maximizing the fitness function,
the population of latent vectors should improve towards the defined requirement through selection and variation. 

At the beginning of each generation, we preserve a copy of the best $N_{elite}$ individuals to save them from alteration. Inheriting them to the next round guarantees that we do not destroy the fittest individuals during a generation.
$N_{pop}$ individuals are selected by conducting $N_{pop}$ tournaments, where
$N_{ts}$ randomly chosen individuals compete against each other based on their fitness.
Two kinds of variation operations, recombination and mutation, are applied to the population resulting from the tournament selection.

\subsubsection{Recombination}
\textit{Uniform crossover} is applied to generate new offspring as in~\cite{bontrager2018deep,fernandes2020evolutionary}. Two latent vectors $a$ and $b$ are recombined to produce two new individuals $\hat{a}$ and $\hat{b}$, with their $i$th attribute randomly chosen from either $a$ or $b$:
\begin{equation}
    \begin{aligned}
   \hat{a}_i = \alpha a_i + (1-\alpha)b_i \text{ and }
   \hat{b}_i = \alpha b_i + (1-\alpha)a_i \text{ with } \alpha=Bernoulli(0.5)
    \end{aligned}
\end{equation}

If recombined, an individual is replaced by its child. 
The crossover rate $p_{cx}$ defines the chance for an individual of the population to participate in recombination. 
As commonly applied, we choose high rates ($0.7$  and $0.9$) to ensure the continuing development of fit individuals~\cite{hassanat2019choosing}.
To introduce new gene material, we add $N_{new}$ new random vectors to the population in every generation in addition to the crossover.

\subsubsection{Mutation}
Following the objective of moving closer to a target cluster, we apply \textit{nonuniform mutation}~\cite{bontrager2018deep}. 
With a probability of 0.5, a variable of a latent vector $z$ is mutated by adding some $noise$ drawn from the original distribution:
\begin{equation}
    \begin{aligned}
       z_i = z_i + noise \sim \mathcal{N}(\mu=0, \sigma=1)
    \end{aligned}
\end{equation}

The mutation rate $p_{mut}$ defines the chance for an individual of the population to be mutated. While the mutation rate of a genetic algorithm is usually set to a few percent~\cite{hassanat2019choosing}, we consider both low ($0.2$) and high ($0.5$) mutation rates, as initial runs showed low diversity among the population.

In accordance with typical tuning methods of evolutionary algorithms, we consider different tournament sizes ($N_{ts} = \{3, 6\}$) and population sizes ($N_{pop} = \{100, 200\}$) adhering to the commonly used parameter choices~\cite{eiben2003introduction,hassanat2019choosing}. The parameters under variation are summarized in Table \ref{tab:expGA}. The number of generations is set to $N_{gen}=500$ as a compromise of running time and complexity of the problem. Table \ref{tab:constantGAparams} displays the constant parameters.
Due to the stochasticity underlying evolutionary systems, some runs naturally never achieve any fitness due to an `unlucky' initialization of the population. Therefore, we test the model for different styles per parameter combination. From a random selection of styles presented to the experimenter, they chose five distinct ones to ensure visual diversity as a basis for the experiments.

\begin{table}[tb]
    \centering
    \caption{GA parameters under variation.}
    \centering
    \begin{tabular}{ p{4cm}>{\raggedleft\arraybackslash}p{2cm}}
    \hline
    Parameter & Setting\\
    \hline
       Crossover rate $p_{cx}$  & 0.7, 0.9\\
       Mutation rate $p_{mut}$ & 0.2, 0.5 \\
       Population size $p_{pop}$ & 100, 200\\
       Tournament size $p_{ts}$ & 3, 6\\
    \hline
    \end{tabular}
    \label{tab:expGA}
\end{table}

\begin{table}[tb]
    \centering
   \caption{Constant GA parameters.}
   \centering
    \begin{tabular}{ p{3.5cm}>{\raggedleft\arraybackslash}p{5.5cm}}
    \hline
    Parameter & Setting\\
    \hline
       Size of individual  &  512\\
       $N_{gen}$  & 500\\
       $N_{elite}$ & 1 \\
       $N_{new}$ & 10 \\
       Recombination operator & uniform crossover ($\alpha=0.5$)\\
       Mutation operator & nonuniform mutation ($\mu=0, \sigma=1$)\\
    \hline
    \end{tabular}
    \label{tab:constantGAparams}
\end{table}

\section{Results}
\begin{figure}[tb]
    \centering
    \includegraphics[height=3.5cm]{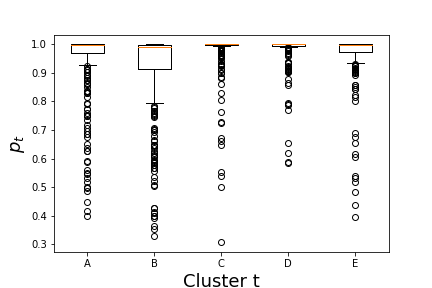}
    \caption{Boxplot of the original images' posterior probability for the five target styles chosen for the experiment. The five clusters range from 228 to 367 in size. The images contained by them have a mean $p_t$ between 0.92 and 0.98.}
    \label{fig:boxplot}
\end{figure}

We tested our model in each parameter combination to generate designs for five different styles.
Fig. \ref{fig:boxplot} displays the distribution of the images' posterior probability per target style cluster.
A comparison of the mean maximum fitness reached across all five runs is presented in Table \ref{genalgres}.
As we ran the whole system for five different times, or style clusters, the results that we include are averages over those five runs. 
Recall that the fitness to be maximized is defined as an image's posterior probability of belonging to a GMM component. Hence, it can vary between a minimum of $0$ and a maximum of $1$. 

\begin{table}[tb]
\centering
\caption[Experimental results.]{Average maximum fitness across all five runs per parameter combination.}
\centering
\begin{tabular}{ll|llll}
\hline
                                          &  & \multicolumn{2}{l}{$p_{mut}=0.2$} & \multicolumn{2}{l}{$p_{mut}=0.5$} \\
                                        &                & $N_{ts}=3$ & $N_{ts}=6$ & $N_{ts}=3$ & $N_{ts}=6$ \\
                                        \hline
\multirow{2}{*}{$p_{cx}=0.7$}     & $N_{pop}=100$ &    0.5746  & 0.2           &     0.6021 & 0.2119  \\
                                        &$N_{pop}=200$ &      0.4031 & 0.5491       &       0.6341 & 0.4028\\
\multirow{2}{*}{$p_{cx}=0.9$}     & $N_{pop}=100$ &        0.4538 & 0.3855         &     \textbf{0.735} & 0.2982 \\
                                        & $N_{pop}=200$ &       \textbf{0.8073} & 0.2       &       0.7043 & 0.5248 \\
\hline
\end{tabular}
 \label{genalgres}
\end{table}
As the comparison of the results reveals, the algorithm finds individuals of highest fitness for a crossover rate of $p_{cx}=0.9$ and a tournament size of $N_{ts}=3$. In particular, the highest fitness is reached in combination with a population size of $N_{pop}=200$ and a low mutation rate of $p_{mut}=0.2$, followed by the second highest fitness with $N_{pop}=100$ and $p_{mut}=0.5$.

\begin{figure}[tb]
    \centering
        \includegraphics[width=9cm]{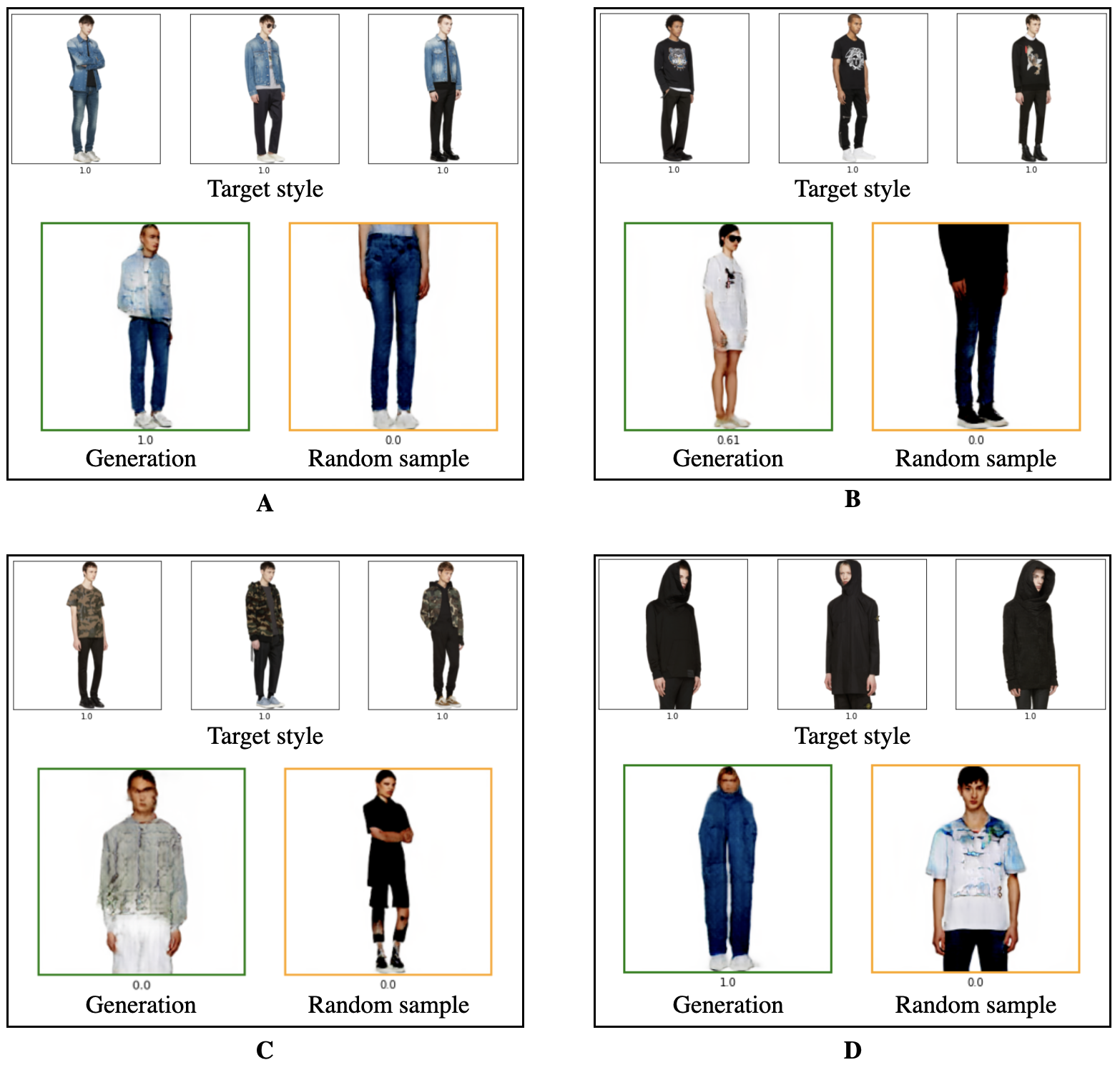}
        
    \caption
    {Four exemplary designs for different clusters. See Fig. \ref{fig:all-plots} for the parameter settings underlying the runs. The figure displays the three images with the highest posterior probability in the style cluster, the fittest generated image, and the fittest out of $N_{pop}\times N_{gen}$ randomly sampled images for comparison. The titles of the images display their fitness. Zoom in for detail.}
    \label{fig:all-designs}
\end{figure}

\begin{figure}[tb]
    \centering
        \includegraphics[width=8cm]{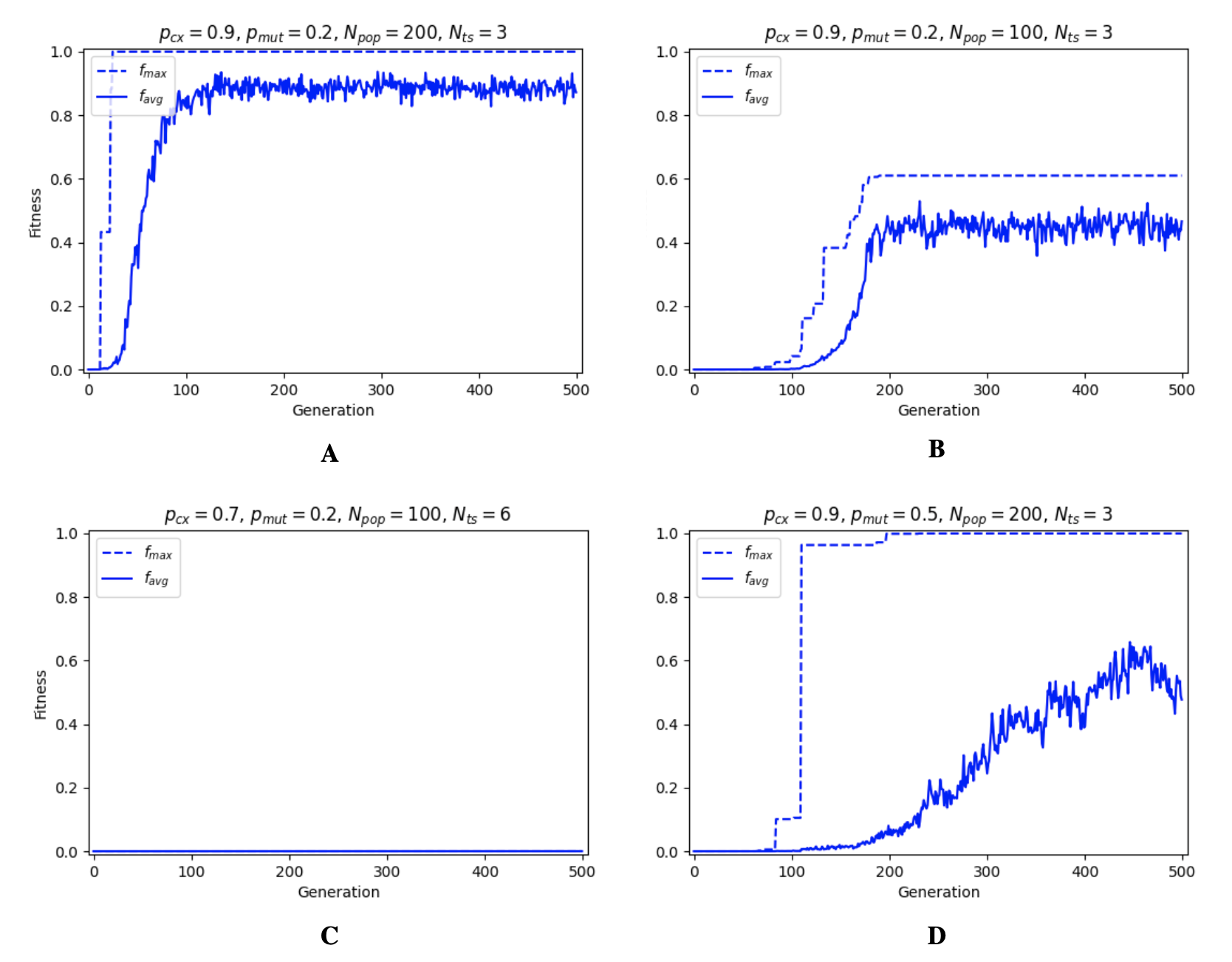}
        
    \caption
    {Maximum and average fitness over generations of the exemplary runs in Fig. \ref{fig:all-designs}.}
    \label{fig:all-plots}
\end{figure}

To better understand the fitness measure in relation to the produced designs, an exemplary case from the experiments for style cluster A, B, C, and D each is presented in Fig. \ref{fig:all-designs}. A plot of average and maximum fitness of each exemplary run is shown in Fig. \ref{fig:all-plots} for inspection of the search behavior. 
For a generated design for the style displayed in Fig. \ref{fig:all-designs}A, the similarity to the target cluster is visible, as they share the features of an open denim jacket with bleached plats and darker pants.
This is aligned with the high fitness of the generation. Interestingly, the random sampling achieved no fitness despite the similarity of the displayed pants.
In comparison, some generated designs of lower fitness, such as the example in Fig. \ref{fig:all-designs}B, show less visual similarity. Specifically, only one feature, namely the pattern on the chest, seems to resemble the target style characterized by a black outfit with a white pattern on the shirt.  

The same concern arises for a generated design for the target style characterized by a camouflaged patterned upper part and black pants, shown in Fig. \ref{fig:all-designs}C. While the image shows an upper body with a pattern similar to the target style, it achieved no fitness, as the corresponding plot in Fig. \ref{fig:all-plots}C shows. While this could be cause by an unlucky initialization, it might also point to flaws in the fitness measure.
%
That the assigned fitness does not seem to be in alignment with visual coherency is also the case for the example given in Fig. \ref{fig:all-designs}D. Here, high fitness is not aligned with a legit outfit design.

\section{Discussion}

For some of the analyzed designs, the fitness measure aligns with the visual coherency to the target style. In Fig. \ref{fig:all-designs}A, high fitness represents a design of high visual similarity. In contrast, lower fitness describes a design of little visual coherency in Fig. \ref{fig:all-designs}B. 
Taking a look at the search behavior behind the latter, displayed in Fig. \ref{fig:all-plots}B, an explanation could be the following. The population converged to a local maximum after around half of the generations, only allowing for optimizing a minor feature in the following generations. 
Though convergence to a local maximum is a desirable end stadium for genetic algorithms, it poses a problem in this scenario. 
Due to weak style matches dominating the population, the subsequent evolution underlies the prevailing look. A solution to introduce more diversity to the population could be to increase mutation or the number of new individuals when fitness stagnates. 

For other examples, the fitness measure does not represent the visual coherency. 
Fig. \ref{fig:all-designs}C displays a design with a pattern similar to the target style, but achieved no fitness. This example raises the question of whether other factors, such as the models' posing, play a role in defining a style. At the same time, the example shown in Fig. \ref{fig:all-designs}D achieved the maximum fitness, for a design that a human judge might not assign validity. 
As our style model is based on the representation of an attribute model trained to recognize clothing features, it might lack the sensitivity of body shapes. 
However, coming with a background in clothing and physiology guides, such notions guide our perception of the designs.
Hence, the suggested system might find some completely different attributes behind what we identify as decisive properties for a style.
We hypothesize that the designs found by the evolutionary search reveal a machine-specific understanding of fashion style.
However, to further investigate that claim, we need to find out whether our model is sufficient to capture styles. As part of our future work, further insight into the implementation and its parameters would show if they optimally support this goal.

The difference in perception could also be because computational networks are not able to capture subtle differences in styles (yet)~\cite{takagi2017makes}. 
Detecting style-coherent similarity of clothing images remains a challenge, in which also the attribute models used to retrieve the visual representation, including the training data, play an essential role~\cite{hsiao2017learning}.
Even though the embedding of images was used instead of the ResNet's final layer's attribute prediction scores, the categories underlying its training influence what the model \textit{sees} on every layer.
To better understand the style-defining features, a qualitative analysis of the ResNet's performance, e.g. with the help of a \textit{Class Activation Map} (CAM), could be performed.
Such an analysis could shed insight on the critical features for a particular fashion style~\cite{takagi2017makes}. With that knowledge, 
the latent space of the GAN might be controllable in a more targeted way.

To quantify the relation between the clusters found by the evolutionary search and human judgement, a human could be added to the loop.  
To guide the system into the direction of human-compatible styles, the approach could eventually be combined with interactive evolution like in DeepIE or StyleIE, in order to integrate how humans see style.
Additionally, we could consider if exploring the latent space can even tell us more about what the model interprets behind styles, providing insight into clustering components. Following that path contributes to understanding how the proposed system sees visual themes emerge, eventually leading to a better understanding of its functioning.

As a final remark, it is essential to reflect on the datasets used for training generative models. As Takagi et al.~\cite{takagi2017makes} point out, fashion photographs do not represent what people actually wear, hence might not give an actual representation of current styles.
Most datasets used for training generative fashion design models consist of catalog or social media images, highly biased to the presented populations. Therefore, a crucial task is to discuss how data affects the system's stability with regard to design diversity, such as achieving desirable silhouettes while still considering diverse body forms.

\section{Conclusion and Future Work}
This research aimed to extend the classical generative deep learning approach to facilitate the generation of designs that respond to fashion styles.
We investigated the application of a genetic algorithm and a GMM to guide the generation of images with regard to previously identified style clusters. Our suggested framework facilitates the search for images responding to certain style clusters.
The experimental results indicate that the proposed procedure provides a promising direction to guide the search for style-coherent designs.
Further research is required to establish a robust and reliable exploration process.

While the system can generate images of maximum fitness, the designs do not necessarily correspond to the target styles in the way one would expect.
We hypothesize that the generations reveal a machine-specific understanding of fashion style.
While some of the generated images exhibit similarity to the target cluster visible to the human eye, other generated outputs raise the question of how the algorithm might understand styles differently, outlining the need to improve the fitness measure. 

Integrating fashion style analysis and fashion generation opens up new design possibilities, such as extending trends forecasting to generating trending designs. 
Future work could investigate how the ability to react to stylistic developments through unsupervised learning expands the capabilities of generative models as creative design tools.




\bibliographystyle{splncs04}
\bibliography{mybib}
%




\end{document}